**Application of the Double Metaphone Algorithm to Amharic Orthography**
***International Conference of Ethiopian Studies XV***
**Daniel Yacob**

**Abstract**
The Metaphone algorithm applies the phonetic encoding of orthographic sequences to simplify words prior to comparison. While Metaphone has been highly successful for the English language, for which it was designed, it may not be applied directly to Ethiopian languages. The paper details how the principles of Metaphone can be applied to Ethiopic script and uses Amharic as a case study. Match results improve as specific considerations are made for Amharic writing practices. Results are shown to improve further when common errors from Amharic input methods are considered.

## Introduction

In the field of text analysis the concept of word distance is introduced and becomes of central importance to the problem. Word distance is the essence of word comparison which appears in innumerable problems of text processing. Text searching and document spell checking are but a few familiar examples.

The notion of distance between groups of written symbols may at first seem highly abstract since we are trained to think of distance as intrinsically spatial. In the natural world we know distance as a geometric orientation between two tangible objects, measured with an equally tangible instrument. Word distances are unitless and determined by the number of corrections required, character by character, to transform one word into another. Each type of correction is given a (likewise unitless) "weight" value where the magnitude may reflect the severity of the discrepancy. The sum of these values then determines the overall magnitude of the word distance.

In naive form all symbols will be treated equally. A set of words of equal length with completely unlike phonemic sequence will appear equally distant from a comparison word. To improve upon the distance distribution, linguistic and orthographical information may be applied as a stage of the distance analysis. Most commonly, linguistic analysis is applied to precondition words to better improve their apparent proximity from the comparison word.

The preconditioning approach is particularly useful and has as its aim the rectification of the disparity between the phonology of the spoken word and the orthography of the written word. For example vowel clusters would be represented by single symbol.

This paper presents an investigation into the application to Amharic orthography of a very successful preconditioning technique, the Metaphone algorithm, used for English and European languages. Before presenting the details of the technique itself, it is useful to first review the nature of the problem that it will be applied to.

## Problems in Amharic Spelling

Amharic orthography reflects the spoken phonetic features to a large extent. So closely that one can be lead to believe that there is no notion of *"spelling"* in Amharic. The rule generally followed is "if a word *sounds* right when read aloud then it was rightly written". Upon closer inspection we quickly realize that Amharic spelling rules are just very forgiving when compared to the strict, albeit irregular, conventions of English. When compared to English, the Amharic author is to a degree liberated to put more cognitive energy into qualitative writing and less into *how* words must be written. Though many fewer and requiring less conscious attention, Amharic spelling very clearly has rules. For example, some phonetical spelling variations are more acceptable than others. While few would give pause to "**ውኃ**" vs "**ዉሀ**", or even "**ታህሳስ**" vs "**ታኅሣሥ**", the rendering "**ዓዲሥ ዐበባ ዒትዮጵያ**" while phonetically valid is *not* an acceptable replacement for "**አዲስ አበባ ኢትዮጵያ**". Though we will explore a number of other categories of Amharic misspellings, spelling error correction may be viewed in large part as a standardization effort.

## Levels of Amharic Spelling

In Ethiopian society there are acceptable levels of precision, a phono-orthographical radius that renderings may fall within to be considered recognizable and acceptable. The sarcastic quip of Mark Twain:

> *"I respect a man who knows how to spell a word more than one way"*

becomes the modus operandi of Amharic spelling. At the most basic level words may be spelt identically as they are spoken. The basic level provides a working medium for all informal exchanges. The basic level is the most pragmatic and so long as character sequences can be converted back into the sounds of the words they record, no judgment is passed over the spelling quality (e.g. "**ባል**", "**መልክት**","**እንቁጣጣሽ**", "**አይን**", "**አሳ**", "**ያገር**"). The glyph forms of letters may also be found to vary here. Persons with minimal education up to high school level will not often venture beyond the basic spelling conventions.

At the intermediate level, more formal word choice is found in diction and variations in word renderings diminish. The writer is likely making a conscious effort to spell with the canonical syllograph from amongst available isophonic alternatives. (e.g. "**በዓል**", "**መልዕክት**", "**እንቍጣጣሽ**", "**ዓይን**", "**አሣ**", "**የአገር**"). The middle level of conformance is expected in novels, government newspapers, and professional literature.

At the most advanced level of writing the canonical forms of words *must* be used. Words can no longer be written merely as they would be spoken. This highest, hypercorrect, level of formal Amharic is largely the realm of specialists. Literature demanding this level of conformance would be Biblical, imperial and dictionaries. Accordingly, these works also become references for canonical Amharic. Writers at the top level have almost invariably had training in Amharic language and literature and usually in Ge'ez as well. (e.g. "**በዐል**", "**መልእክት**", "**ዕንቍጣጣሽ**", "**ዐይን**", "**ዐሣ**", "**የሀገር**").

Level boundaries are by no means discrete, rather they are blurred (particularly in the lower levels) and practices normally found at a higher level may in fact be found mixed within a manuscript that predominately exhibits features of a lower level. As might be expected, convention drift occurs more in the downward direction than upward. An Egyptian or Aztec pyramid model is somewhat useful to consider. What these structures help illustrate are the wider degree of spelling variance considered acceptable at the base and greater conformance towards the canonical at the peak. To further define the three level model and debate features such as shape, thickness, and sublevels is of limited practical value for the problem of word rendering validation. The three levels are not formally recognized and offer no more than a heuristic model.

Arguments can be made for software to support the spelling conventions of the top two levels. Amharic spelling software should always support the canonical renderings. Though in a number of cases correctly rendered words may not be immediately recognized by the average person and will appear out of place in the informal manuscript. Time will tell. User feedback and perhaps a market survey would be required to determine if the software support for pencanonical conventions would be worth developing.

**Errors Inherent from Symbol Redundancy**

As it is well known, much of Amharic's vocabulary has its origins in the Ge'ez language. Amharic maintains the entirety of Ge'ez symbology while preserving most, but not all of, Ge'ez's phonology. As a consequence, the phonemes of a number of symbols in the orthography have decayed into the nearest Amharic equivalent. The resulting redundancy in symbols in Amharic is easily the most prevalent and readily observable influence that complicates Amharic spelling. For the purpose of future reference we will label this issue as "Type 1: Syllographic Redundancy".

While the symbols are homophonic equivalents they are not interchangeable at the top level. At the lower levels, while the homophonic characters may be interchanged freely not all substitutions that are possible are also probable. Nonetheless, an Amharic spellchecker should be capable of equating all homophonic equivalents and be able to offer the canonical form.

**Table 1: Syllographic Redundancies**

| Canonical Amharic | Common Amharic | Possible but Improbable Amharic |
|---|---|---|
| ዓለም | አለም | ዐለም, ኣለም |
| ፀሐይ | ጸሃይ, ፀሃይ, ጸሐይ | ጸሀይ, ጸሓይ, ጸኀይ, ጸኃይ, ጸኻይ, ፀሀይ, ፀሓይ, ፀኀይ, ፀኃይ, ፀኻይ |
| ኃይለ ሥላሴ | ኀይለ ሥላሴ[7], ኃይለ ስላሴ, ኀይለ ስላሴ, ኀይለ ስላሤ, ኃይለ ስላሤ, ሀይለ ስላሴ, ሀይለ ስላሤ, ሀይለ ሥላሴ, ሃይለ ስላሴ, ሃይለ ሥላሴ, ሃይለ ስላሤ, ኀይለ ሥላሤ, ኃይለ ሥላሤ, ሀይለ ሥላሤ, ሃይለ ሥላሤ, ሐይለ ስላሴ, ሐይለ ስላሤ, ሐይለ ሥላሴ | ሐይለ ሥላሤ, ሓይለ ሥላሤ, ሓይለ ሥላሴ, ሓይለ ስላሤ, ሓይለ ስላሴ, ኻይለ ሥላሤ, ኻይለ ሥላሴ, ኻይለ ስላሤ, ኻይለ ስላሴ |

In addition to the phonetic redundancy of characters, Amharic suffers slightly from visual redundancy in a few cases. Most prominently the vowel markers of; ‘ው’ and ‘ዉ’ , ‘ፕ’ and ‘ኘ’, ‘ፖ’ and ‘ፓ’, ‘ዩ’ and ‘ዪ’, ‘ጉ’ and ‘ጕ’ are similar enough that the former characters are often interchanged in words with the later. This problem is exacerbated at small print sizes and with the lack of visual clarity often found on computer screens. Additionally, the letter ‘ቁ’ is often used in place of ‘ቍ’ (e.g. “ቍጥር” vs “ቁጥር”) which may owe more to phonetic proximity and decay than to visual (again type face quality plays a role here). Often times the writer may simply be choosing the form that is easiest to write by hand or type into a computer.

A similar problem that stems from confusion with Ethiopic vowel markers occurs with the arm-and-ring vowel marker used for both 7th and 8th form syllables. The arm-and-ring marker makes its first and most frequent appearance on the 7th syllable ‘ሎ’ for which the sign is then most associated. The arm-and-ring-marker shows up again with the infrequently used 8th form elements ‘ቈ’, ‘ኈ’, ‘ኰ’ and ‘ጐ’ in the Ge’ez character set. Given the infrequent and somewhat specialized use of these rounded labial elements, their purpose is often forgotten by the layman and they become used as alternatives to their 7th form counter parts. Note that ‘ጐ’ and ‘ኰ’ are the most often substituted letters, and that the rules reverse for ‘ቈ’ which becomes the substitutee, though substitutions may occur in both directions for all elements.

**Table 2: Glypheme Misidentifications**

| Canonical Amharic | Common Amharic |
|---|---|
| ቈረሰ | ቆረሰ |
| መኰንን | መኮንን |
| ጎንዳር | ጐንዳር |

It is noteworthy that the arm-and-ring also gets added to ‘የ’ as an alternative rendering of the 7th member of the ‘የ’ family. Also deserving of attention are letter face design styles whereby ‘ጬ’ and ‘ጮ’ will appear as ‘ጨ’ and ‘ጬ’ respectively. Occasionally an author who is more familiar with the later style may out of habit apply ‘ጬ’ in place of ‘ጮ’ when it is inappropriate for the working typeface. We will later refer to this issue of letter appearance problems as “Type 2: Glypheme Misidentification”.

**Errors Inherent from Phonology-Orthography Disconnect**

As another consequence of the decline of Ge’ez phonemes in Amharic discussed earlier, Ge’ez words in Amharic may be rendered identically or may have lost one or more letters over time. In some cases letters that do remain in canonical Amharic rendering have become silent. Having become superfluous, the silent letters are commonly dropped or elided at the basic and intermediate levels. A writer working at the intermediate level may wish to use the canonical rendering but insert the wrong letter. The Ge’ez style pluralization is another example of a “false Ge’ezism” that may occur when a word is thought to be of Ge’ez origin or maintain Ge’ez rules. We will label this issue “Type 3: False Ge’ezisms”. In some instances, mostly in older materials that were not electronically produced, a false Ge’ezism may in fact be an archaic Amharic spelling:

**Table 3: False Ge’ezisms**

| Canonical Amharic | False Ge’ezism / Archaic Amharic | Ge’ez |
|---|---|---|
| ምልክት | ምልእክት | ምልክት[7] |
| ቀለሞች | ቀለማት | ቀለማት |
| ትዕግሥት | ትዕግዕሥት | ትዕግሥት |
| አየ[15] | ዐየ[15] | ርዕየ |
| ሁሉ | ኍሉ, ኹሉ[6] | ኵሉ[7] |
| ሁለት | ኍለት, ኹለት[6] | ክሌቱ[7] |

There are a number of common cases where phonology clashes with Amharic orthography. For instance, ‘ም’ may be exchanged for ‘ን’ before ‘በ’, as in “አንበሳ” vs “አምበሳ”. Likewise ‘ም’ may also replace ‘ን’ before forms in ‘ፈ’ (e.g. “ላንፋ” vs “ላምፋ”).[15] Spoken Amharic has a great many alternations, whole and partial assimilations. Not all spoken occurrences will also manifest themselves in written form. For the sake of simplicity and efficiency we attempt to account only for occurrences encountered in surveyed literature. “Type 4: Assimilations and Alternations”.

The occurrence of the morpheme –(i)ያ- is a source of much confusion. This is no doubt due to the similarity between the phonology of collective nouns and the noun of instrument. The nouns of instrument can be considered a derivative of the infinitive where the suffix "+iያ" is appended causing the ending 6th form syllable to transform into the 3rd (e.g. "መስበር" ⇒ "መስበሪያ").[15] Some collective nouns work the opposite way where the 3rd from will transform into the 6th as the noun of agent of agent is pluralized with -iያን (e.g. "አንባቢ" ⇒ "አንባብያን"). "ክርስቲያን"[3] is a special case that breaks the collective noun rule:

**Table 4: Phonology-Orthography Disconnects in –(i)ያ-**

| Canonical Amharic | Common Amharic |
|---|---|
| አንባብያን | አንባቢያን |
| ወንጌላውያን | ወንጌላዊያን |
| አብዮታውያን | አብዮታዊያን |
| ኢትዮጵያውያን | ኢትዮጵያዊያን |

Though it has not been stated until now, our focus is on underived words. Any and all affixes must be stemmed before the Metaphone algorithm can be applied. The complexity with –(i)ያ- will also impact the rendering of underived nouns that feature a natural 6th form syllable plus 'ያ'. In such cases the 6th form syllable may be rendered in the 3rd form:

**Table 5: Additional Phonology-Orthography Disconnects in -iያ-**

| Canonical Amharic | Common Amharic |
|---|---|
| ሚያዝያ | ሚያዚያ |
| ማርያም | ማሪያም |
| ሰማንያ | ሰማኒያ |

Phoneme-morpheme boundaries are generally preserved in canonical Amharic but may be lost in prefix elisions in the orthography (e.g. "ያገር" vs "የአገር", "ያማርኛ" vs "የአማርኛ", "በንደ" vs "በእንደ"). Elisions may also occur when an alef-vowel follows a 6th form syllable "መልአኩ" ⇒ "መላኩ". The elision occurrences we will label "Type 5: Orthographic Abbreviations and Elisions".

Labio-velarizations present a problem similar to the -iያ- morpheme. It is not readily predictable when the 'ዋ' inflection should join with the preceding syllable or remain separate (e.g. "እሷ" vs "እስዋ"). The confusion with labio-velar renderings we will label now "Type 6: Disjoint Labiovelars":

**Table 6: Disjoint Labiovelars**

| Canonical Amharic | Common Amharic |
|---|---|
| ሆኗል | ሆኖዋል, ሆኖአል |
| ጡዋት | ጥዋት, ጠዋት, ጧት |
| ይዟል | ይዞአል, ይዞዋል, ይዝዋል |

Regional dialects can also impact word formation in the basic level where the words are more likely to be written following their spoken form; "ሂጂ" vs "ሂጅ", "አይዶለም" vs "አይደለም", "ዓጤ" vs "ዓፄ".

An Amharic spellchecker should be able to associate the canonical, common and commonly misrendered forms together. Regional and dialect variations that impact spelling we will label "Type 7: Dialect Variations".

**Errors Inherent from Foreign Language Transcription**

When words arrive to Amharic from outside languages, occurring very often for technology terms, the assimilation process requires at least 3-5 years. Until a stable convention emerges, contrasting transcriptions will be found in popular literature. An Amharic spellchecker should be able to recognize all candidates as related and later present the favored convention once identified. The conflicting transcriptions problem we will now label "Type 8: Foreign Language Transcription":

**Table 7: Foreign Language Transcriptions**

| Canonical Amharic | Common Amharic |
|---|---|
| ቴክኖሎጂ[3] | ቴክኒዎሎጂ[16] |
| ኮምፒዩተር[3] | ኮምፒውተር[16] |
| ኢሜይል[3] | ኢሜል, ኤሜል, ኤሜይል |

**Errors Inherent from the Typing System**

Every means of lettering, from a stone stylus to a PDA stylus has its own innate advantages and hazards. Keyboard composition is the most widely used form of rapid personal character rendering in the present day. Keyboards offer faster character rendering by removing from the writers hand the burden of forming the shape of each letter. The keyboard offers an interface to a mechanical or electrical apparatus that will create the shape of each letter more quickly and consistently than a human ever could. The writer need only tap the symbol of the desired character on a keyboard "key".

With proficiency, the writer will use all fingers to strike keys independently and without observation or even thought of what the hands are doing. Word formation errors that are likely from keyboard composition are related to the geometric distribution of the letter symbols, the skill of the writer, and the complexity of the mechanical motions required of the hands to compose a letter.

As the number of keys on a keyboard is invariably fewer than the number of symbols of the Ethiopic writing system; Ethiopic input methods must be defined such that the majority of letters will require that several keys be struck for composition. The possibility exists in the course of composition that the writer will omit a key or strike a key other than the intended (a "mistrike"). Left uncorrected these mistrikes will lead to a word formation error.

The type of mistrikes that may occur depends highly on the input method. Mistrikes that are common under one input method (such as the Ethiopian typewriter) may be less likely or impossible with another (such as phonetic based). Independent from the keyboard itself, the rendering apparatus may not make available a desired letter or symbol. In these cases the writer may elect to substitute the unavailable letter with another, omit it, or compose a facsimile of the symbol thru other available symbols and glyphemes. We will apply the common name here, "mistrike", to identify errors of this class as "Type 9: Mistrikes".

To this end we consider the common mistrikes inherent to phonetic based input methods. The most familiar example mistrikes from phonetic based input methods come from the "shift-slip" condition. Whereby a typist's fingers have pressed or released the shift key either too soon or not soon enough resulting in the wrong letter being composed.

**Table 8: Mistrikes of Phonetic Input Methods**

| | | |
|---|---|---|
| **Shift-Slip** | ሐቐኘኸዽጝጠጨጸጰዠ | ሀቀነከደግተቸሰጸዘ |
| **5th form for the 1st** | ጤና | ጠና |
| **Unintended Elision** | አርኢያ | አሪያ |

**Metaphone and Sound Based Word Comparison**

The Soundex (or Miracode), Metaphone and Double Metaphone algorithms belong to the family usually known as "phonetic encoding" or "sound alike" algorithms that offer a heuristic type of fuzzy matching. The algorithms were developed to address the problem of indexing personal names in English which often come from non-English languages and consequently result in a variety of renderings. The Soundex algorithm was devised in 1935 to index the US Census results of 1880-1920. The Metaphone algorithm was introduced more than half a century later in 1990 and offers considerable improvements over its predecessor. These algorithms take the approach of converting a word into a simplified form based on the phonetic patterns that the letter sequences represent. The resulting simplified words will then appear "nearer" to one another in written form.[8,13]

In essence these name matching algorithms attempt to untangle the rules of English orthography so that names like "Steven", "Stephen" and "Stefan" will be recognized as spelling variations of the same name. Variations on personal name spellings abound in English (e.g. Nelson, Nielson, Niellson, Nelsonn, Nelsone, Nellson, Nielsone, Nielsen, etc) and while none are deemed "wrong" it is useful to be able to associate them together. In particular the field of genealogy relies heavily on these algorithms for just this purpose.

Soundex and its decedents are still very much used in the present day. While designed as a tool to undo the problems inherent with English transcriptions of proper nouns, the algorithms are applied to every part of speech. With the rise of word processors and desktop publishing the algorithms have a new found and vital role at the heart of spelling correction software (aka "spellcheckers"). Under this new utilization the algorithms are applied not to match variations of valid spellings but rather to find a valid word spelling when presented with one known to be invalid.

**Metaphone for Amharic**

The Metaphone algorithm, like the other members of its family, exists to serve as tools to operate on English words and in the English alphabet. Without modifying the Metaphone algorithm, it could not be applied to Amharic unless Amharic words were first converted into a representation under the English alphabet. Doing so we would anticipate a lower success rate for Amharic than we find for English as rules of English orthography are applied out of context on symbol sequences that do not represent English words. Such an approach would however be useful as a basis of comparison for a Metaphone algorithm derived specifically for Amharic.

To develop a Metaphone algorithm for Amharic we will want to take into consideration the problems that we have identified earlier that are particular to Amharic spelling. Firstly, we will assume throughout that we are working in the Ethiopic character set. Next, we would want our algorithm to be insensitive to the distinctions between symbols of differing appearance but of like sound. Somewhat analogous to how the Metaphone algorithm for English ignores letter case by converting words into a single case, we want to convert phonetically redundant letters into a single representative. Thus we have the following conversion rules:

**Table 9: Simplification of Phonetically Equivalent Syllables**

| Phonemic Equivalents | Simplification |
|---|---|
| ሀ, ሃ, ሐ, ሓ, ኀ, ኃ, ኻ<br>ሑ, ኁ, ኹ<br>⋮ ⋮ ⋮ | ሀ<br>ሁ<br>⋮ |
| ሰ, ሠ<br>ሱ, ሡ<br>⋮ ⋮ | ሰ<br>ሱ<br>⋮ |
| አ, ኣ, ዐ, ዓ<br>ኡ, ዑ<br>⋮ ⋮ | አ<br>ኡ<br>⋮ |
| ጸ, ፀ<br>ጹ, ፁ<br>⋮ ⋮ | ጸ<br>ጹ<br>⋮ |
| ቁ, ቍ | ቍ |
| ቆ, ቈ | ቈ |
| ኮ, ኰ | ኮ |
| ጎ, ጐ | ጎ |

An example word "ዓለምፀሐይ" would become "አለምጸሀይ" under these conversions. In so doing we overcome the Type 1 problem. The next step that we will perform will be to apply the Metaphone rule of removing all vowels from a word except an initial vowel should a word start with a vowel. The elimination of vowels is the most heavy-handed step in the Metaphone process, but it is in large part the key to its success. Eliminating the vowels in Amharic words takes care of the bulk of the matching issues identified earlier. Vowels are eliminated by removing all Alef-A (አ) and Aynu-A (ዐ) forms, and then by converting all syllables into the Sadis, or $6^{th}$, form. Applied to our example word, "አለምጸሀይ" simplifies further to "እልምጽህይ". In this step we overcome the identified problems of Type 2, 4 and 5 as well as some occurrences of Types 7, 8 and 9.

In this simplification process we do make two special exceptions. The first is to preserve 'ኘ' in its Ge'ez form which we will want to compare to 'ፕ' shortly to address the glyph mismatch problem. The other exception is to convert the labiovelar forms of the fourth order into two Sadis letters. The first would be the Sadis form of the syllable itself, followed then by a Sadis ‘ው’. Doing so allows us to compare renderings for the Type 6 disjoint labiovelar problem. For example; ‘ኳ’ reduces to “ክው” and the remaining labializations of ‘ከ’ (‘ኰ’, ‘ኲ’, ‘ኵ’ and ‘ኴ’) ultimately become ‘ክ’ as no examples of Type 6 problem have been found for these forms. We continue on to handle spelling mistakes that arise from glyph confusion, common typing errors, and phonology issues.

Under the Double Metaphone algorithm, a follow up to the Metaphone algorithm arriving in the year 2000, two simplified phonetic encodings of a word can be generated. The first encoded word is considered the most probable as a representation for the target word. When appropriate, a second encoded word is created that offers an alternative encoding under less frequently used English conventions. For Amharic this approach remains useful and we will apply it as alluded to earlier. However, for the purposes of investigation we do not want to restrict the possible outcomes to only two encodings. Rather, we will develop as many encodings as appear to be useful under Amharic writing conventions.

The letter ‘ኘ’ presents a case where we would want to encode the letter canonically as ‘ኝ’ and alternatively as ‘ፕ’ for cases where the letter identities were mistaken (we only need do this for the first form of ‘ኘ’ as the mistaken identity problem does not occur with the other syllabic forms). Similarly we will require an additional encoding of ‘ዽ’ for ‘ጵ’ as part of our treatment for the Type 2 problem. Unfortunately these measures do not go far enough and ‘ኘ’ and ‘ዽ’ could have both been results of the shift-slip condition of Type 9 problems. Yet another set of encodings will be created for this possibility which requires an encoding of ‘ጵ’ for ‘ዽ’ and ‘ን’ for ‘ኝ’. The full set of shift-slip substitutions are in the following table, note that the substitutions occur in both directions (uppercase-for-lowercase and lowercase-for-uppercase):

**Table 10: Mistrikes of Phonetic Input Method**

| Lowercase | Uppercase |
|---|---|
| ስ | ጽ |
| ቅ | ቕ |
| ት | ጥ |
| ች | ጭ |
| ን | ኝ |
| ክ | ኽ |
| ዝ | ዥ |
| ግ | ጝ |
| ፕ | ጵ |

Operating on our example word for case mismatch issues we would create the second encoded word “እልምስህይ” which we will carry along with “እልምጽህይ” for all future operations. The step of checking for case mismatches, which addresses problem Type 9, is worthy of further discussion. It must first be noted that doing so is closely tied to an assumed input method. The table here presents shift variants that are applicable to many popular phonetically based input methods in use presently. However, the table is totally inappropriate for analyzing text that would have been keyed using an input method based on the Ethiopian typewriter for example. An Ethiopian typewriter based input method has a completely different set of error types associated with it and would require a different corrective procedure at this stage. The “keyboard distance” between words should always follow a Metaphone procedure. However, given the special and high frequency nature of the issue for Ethiopic input methods, the shift-slip condition is analyzed at this stage. Accordingly, this step should be used with care and when the input method associated with the document is unknown, should not be used at all. The alternative encodings from this step are given the lowest priority.

The Metaphone algorithms for English will encode some phonetically similar consonants together. Such as the labials ‘b’ and ‘p’ both encode as ‘b’, the dentals ‘t’ and ‘d’ both become ‘t’.[8] This encoding strategy may be advantageous for Amharic as well. Until a much larger and more thorough investigation can be conducted to determine the merits of doing so, we cautiously encode Amharic consonants separately. A single exception is made at this time for the encoding of ‘ቭ’ as ‘ብ’ which should not significantly impact our smaller exploratory investigation as very few Amharic words will use the ‘ቨ’ family, and then only for transcription of foreign words.

Finally, the only remaining phonological issue we have to address at this time is the substitution of ‘**ም**’ for ‘**ን**’ before ‘**ብ**’ and ‘**ፉ**’. Encoded words are created for each.

To help illustrate the algorithm steps we have just discussed in prose, the following table presents encoding examples for misspellings of “**ዓለምፀሐይ**”, “**ጡዋት**”, “**ወንበር**”, “**ፕሬዚደንት**” and the correctly spelled “**ላም**”:

**Table 11: Steps in Amharic Metaphone**

| **Metaphone Step** | **አለምጸሐይ** | **ጧት** | **ወምበር** | **ኘሬዚዳንት** | **ላም** |
|---|---|---|---|---|---|
| 1) Simplify | **አለምጸሀይ** | **ጥዋት** | **ወምበር** | **ኘሬዚዳንት** | **ላም** |
| 2) Remove Vowels | **እልምጽህይ** | **ጥውት** | **ውምብር** | **ኘርዝድንት** | **ልም** |
| 3) Check for Phonological Problems | **እልምጽህይ** | **ጥውት** | **ውምብር**<br>**ውንብር** | **ኘርዝድንት** | **ልም** |
| 4) Check for Glyph Mismatches | **እልምጽህይ** | **ጥውት** | **ውምብር**<br>**ውንብር** | **ኝርዝድንት**<br>**ፕርዝድንት** | **ልም** |
| 5) Check for Input Method Problems | **እልምጽህይ**<br>**እልምስህይ** | **ጥውት**<br>**ጥውጥ**<br>**ትውጥ**<br>**ትውት** | **ውምብር**<br>**ውንብር** | **ኝርዝድንት ኝርዝዽንት**<br>**ፕርዝድንት ፕርዝዽንት**<br>**ንርዝድንት ንርዝዽንት**<br>**ኝርዥድንት ኝርዥዽንት**<br>**ንርዥድንት ንርዥዽንት**<br>**ፕርዥድንት ፕርዥዽንት**<br>**ኝርዝድኝት ኝርዝዽኝት**<br>**ንርዝድኝት ንርዝዽኝት**<br>**ፕርዝድኝት ፕርዝዽኝት**<br>**ኝርዥድኝት ኝርዥዽኝት**<br>**ንርዥድኝት ንርዥዽኝት**<br>**ፕርዥድኝት ፕርዥዽኝት**<br>**ኝርዝድንጥ ኝርዝዽንጥ**<br>**ንርዝድንጥ ንርዝዽንጥ**<br>**ፕርዝድንጥ ፕርዝዽንጥ**<br>**ኝርዥድንጥ ኝርዥዽንጥ**<br>**ንርዥድንጥ ንርዥዽንጥ**<br>**ፕርዥድንጥ ፕርዥዽንጥ**<br>**ኝርዝድኝጥ ኝርዝዽኝጥ**<br>**ንርዝድኝጥ ንርዝዽኝጥ**<br>**ፕርዝድኝጥ ፕርዝዽኝጥ**<br>**ኝርዥድኝጥ ኝርዥዽኝጥ**<br>**ንርዥድኝጥ ንርዥዽኝጥ**<br>**ፕርዥድኝጥ ፕርዥዽኝጥ** | **ልም** |

The example with the Amharic transcription of “**ፕሬዚዳንት**” helps illustrate a problem in encoded word generation that occurs in the fifth step for input method correction. Most plainly the approach leads to the possibility of a high number of encodings being generated. The occurrence of more than one input method related error per word is not frequent. The encodings that are generated in step five represent all possible errors, very few of which are probable and worth the computational cost of further investigation.

We can reduce the number of encodings generated by restricting only one input method error to appear per encoding. This would give us three encodings total for “**ጧት**” and eleven encodings for “**ኘሬዚዳንት**” which is still too high. The number of input method error encodings can be reduced even further by using a single “lowest common denominator” encoding that itself represents an encoding that is a reducible form of all other input method error encodings. With our example input method system, such an encoding presents itself when we apply the shift-slip condition to each key and in the uppercase to lowercase direction only. In Step 5 we would then generate:

**Table 12: Amharic Metaphone Adjusted Step 5**

| Metaphone Step | አለምጸሐይ | ጧት | ወምበር | ፕሬዚዳንት | ላም |
|---|---|---|---|---|---|
| 5) Check for Input Method Problems | እልምጽህይ<br>እልምስህይ | ጥውት<br>ትውት | ውምብር<br>ውንብር | ፕርዝድንት<br>ፐርዝድንት<br>ንርዝድንት | ልም |

The properly spelled word "ላም" has been used as a simple example to help demonstrate and further explain how the matching process works. "ላም" is an easy choice to work with as it is only two characters in length and was not subject to additional encodings as created in steps 3-5. "ልም" is the only possible encoding for "ላም". "ልም" however is a possible encoding for some 48 additional letter sequences. The 49 sequences would come from the first seven forms of "ለ" permuted against the first seven forms of "መ". "ላም" is then one of these 49 permutations that we know to be a valid word. The other 48 permutations would be invalid renderings of "ላም" specifically but do not necessarily represent invalid words in and of themselves. "ሎሚ" for example is a valid word in the remaining set of 48 though not a valid rendering of "ላም":

**Table 13: Valid Alternatives for "ሊም"**

| Invalid Word | Error Encoding | Valid Matches for Error Encoding (With Inflections) | Valid Matches for Error Encoding (No Inflections) |
|---|---|---|---|
| ሊም | ልም | ላም, ላሙ, ላሜ<br>ለም, ለሙ, ለማ<br>ልም, ልሙ, ልሚ, ልማ<br>ሎሚ | ላም<br>ለም, ለማ<br>ልም, ልሚ, ልማ<br>ሎሚ |

It is useful to reflect at this point on the role that the Metaphone algorithm plays in the process of spelling correction. When a word, such as "ሊም" is not found in an electronic "dictionary" we have identified it as an unknown word or an incorrect spelling of a known word. The Metaphone algorithm is then used to simplify the error word for comparison against the simplifications of known words from the dictionary. The simplification of "ላም", "ሎሚ" and a few others will be found to "match" (have zero distance from) the simplification of the error "ሊም". These matching words become candidates for the correction. Typically the list of candidates will then be presented to the typist for selection and substitution for the error word ("ሊም"). The candidates will be presented in an order of greatest probability for the intended word. The proximity distance from the error word (is "ላም" or "ሎሚ" closer to "ሊም"?) is a major factor in determining the probability.

With this understanding, Table 14 presents matches that the algorithm will find for two of our example words that have numerous common misrenderings.

**Table 14: Matchable Likely Misspellings of "ዓለምፀሐይ" and "ጡዋት"**

| Matchable Misspelling of ዓለምፀሐይ (All Variations Encode to እልምጽህይ) | Matchable Misspelling of ጡዋት (All Variations Encode to ጥውት) |
|---|---|
| ዓለምፀሃይ, ዓለምፀሀይ,ዓለምጸሐይ, ዓለምጸሃይ, ዓለምጸሀይ, ዐለምፀሐይ, ዐለምፀሃይ, ዐለምፀሀይ, ዐለምጸሐይ, ዐለምጸሃይ, ዐለምጸሀይ, አለምፀሐይ, አለምፀሃይ, አለምፀሀይ, አለምጸሐይ, አለምጸሃይ, አለምጸሀይ | ጥዋት, ጠዋት, ጧት |

**Experimental Results**

A test of the algorithm procedures described here was conduction on 116 canonically spelled words with 166 corresponding misspellings. The test sample represents the nine categories of misspelling presented in this paper. The Amharic Metaphone algorithm was compared against the Double Metaphone algorithm for the same test set.

The collection of sample words was first transliterated into an English representation so that the Double Metaphone algorithm could then be applied. The semantics of a transliteration system could have a significant impact on the success of Double Metaphone. For instance, the degree to which a transliteration system relied on non-alphabetic elements (punctuation, numerals) could make it more or less amenable to the algorithm. To counter this effect, the Double Metaphone algorithm test was then conducted on a total of four different transliteration systems:

**Table 15: Word Matching Results for 166 Misspelled Forms of 116 Words**

| Procedure | Matches | Percentage |
|---|---|---|
| Ethiopic Script with Amharic Metaphone[2] | 150/166 | 90% |
| SERA Transliteration with Double Metaphone[9] | 105/166 | 63% |
| Ethiop Transliteration with Double Metaphone[9] | 99/166 | 60% |
| ISO/TC46/SC2 Transliteration with Double Metaphone [9] | 97/166 | 58% |
| Mainz Transliteration with Double Metaphone[9] | 88/166 | 53% |

The results are clearly most favorable to the Amharic Metaphone algorithm where words were matched 90% of the time. The discrepancy between the Ethiop, ISO, and SERA transliteration systems is negligible and might fluctuate with the size of the test set. At comparable levels these transliteration systems rely on punctuation to denote letter symbols. While the Mainz system avoids the use of punctuation for letter symbols altogether, it was likely more susceptible to the English orthography rules applied by Double Metaphone and consequentially found the lowest score.

The Amharic Metaphone results were later improved by applying the standard Metaphone rule of treating 'w' and 'y' under vowels rules. With the analogous letters '**ወ**' and '**የ**' treated likewise as vowels an additional 10 words were matched with their misspellings from the test set bringing the Amharic Metaphone success rate up to 96%.

While the test set was representative of our nine misspelling categories not all misspellings fell under the scope of Metaphone's applicability. A perfect matching score was not anticipated but conducting the test helps to illuminate the limitations of the approach that follow on distance calculation algorithms will have to address. In particular the Amharic Metaphone algorithm had difficulty matching Type 7 words when the consonant component of a syllable would change.

**Conclusion & Further Research**

It is evident from the limited test conducted that higher matching rates are found for Amharic words when the Metaphone algorithm is adjusted to apply the rules of Amharic orthography in lieu of those of English. This positive result has held up as the test set of words has grown in number, a trend that is expected to continue. However, a much more thorough test with a larger corpus of materials should certainly be conducted. The development here has been no more than exploratory in nature.

Further research should investigate the benefits of; applying the vowel rules to the '**ወ**' and '**የ**' families, encoding the '**ጠ**', '**ጸ**' and '**ፀ**' families together, encoding '**ፐ**' and '**ፈ**' with '**በ**' and possibly '**ጀ**' with '**ዠ**'[11]. Additional alternations in writing should be investigated as well as mistrikes that occur from well established input methods. With a large corpus of materials an over matching (false-positives) analysis should be made to evaluate the merits of these suggestions and to further optimize the algorithm.

Ultimately any approach to Amharic spelling correction is limited by the reliability of the reference used for canonical formations. The establishment of a comprehensive and authoritative lexicon for written Amharic would be the single most valuable resource towards the realization of this eventual goal.

## Appendix

### Type 1: Syllographic Redundancy

| Canonical Amharic | Common Amharic |
|---|---|
| ሆሣዕና | ሆሳእና, ሆሳና, ሆሣና |
| ሐምሌ | ሀምሌ, ሃምሌ, ኃምሌ |
| ሒሳብ | ሂሣብ |
| ሕግ | ህግ |
| ሥላሴ | ስላሴ, ስላሤ, ሥላሤ |
| ሥጋ | ስጋ |
| ታኅሣሥ | ታህሳስ |
| ኀምሳ | ኃምሳ, ሃምሳ |
| ኃይለ | ኀይለ, ሀይለ, ሃይለ, ሐይለ |
| ድኅና | ደህና |
| ዐወቀ | አወቀ |
| ዓለምፀሐይ | ዓለምፀሃይ, ዓለምፀሀይ,ዓለምጸሐይ, ዓለምጸሃይ, ዓለምጸሀይ, ዐለምፀሐይ, ዐለምፀሃይ, ዐለምፀሀይ, ዐለምጸሐይ, ዐለምጸሃይ, ዐለምጸሀይ, አለምፀሐይ, አለምፀሃይ, አለምፀሀይ, አለምጸሐይ, አለምጸሃይ, አለምጸሀይ |
| ዓሣ | አሳ, ዓሣ |
| ዕንቊጣጣሽ | እንቁጣጣሽ |
| ወሀ | ውኃ, ዉሃ |

### Type 4: Assimilations and Alternations

| Canonical Amharic | Common Amharic |
|---|---|
| ሀገር | አገር |
| ላንፋ | ላምፋ |
| ሽንብራ | ሽምብራ |
| ብሎአቸው | ብሎዋቸው |
| ቅርንፉድ | ቅርምፉድ |
| ተባዕት | ተባት |
| ኀምሳ | አምሳ |
| አንበሣ | አምበሳ |
| አንበጣ | አምበጣ |
| እንቢ | እምቢ |
| እንብርት | እምብርት |
| አንፋር | አምፋር |
| ከበጉዋይ | ከበኋይ, ከበጐይ |
| ውሽንፍር | ውሽምፍር |
| ወንበር | ወምበር |
| ዝንብ | ዝምብ |
| ግልንቢጥ | ግልምቢጥ |
| ግንብ | ግምብ |
| ግንፎ | ግምፎ |
| ጥንብ | ጥምብ |

### Type 2: Glypheme Misidentification

| Canonical Amharic | Common Amharic |
|---|---|
| ሴንሰርሺፕ | ሴንሰርሺኘ |
| ቈነሰ | ቆነሰ |
| ቈይ | ቆይ |
| ቈጠረ | ቀጠረ, ቆጠረ |
| ቍጥራቸውም | ቁጥራቸውም |
| ቍረጥ | ቁረጥ |
| ቍርባን | ቁርባን |
| ቍር | ቁር |
| ቍርስ | ቁርስ |
| ብርኰት | ብርኮት |
| ነኍላላ | ነሁላላ |
| ነው | ነዉ |
| አውሮፕላን | አውሮኘላን |
| ኢትዮጵያ | ኢትዮዽያ |
| ኮከብ | ኰከብ |
| ኮኦፕሬሽን | ኮኦኘሬሽን |
| ኰነነ | ኮነነ |
| ኵላብ | ኩላብ |
| ደቈስ | ደቆስ |
| ዲፕሎማት | ዲኘሎማት |
| ዲፕሎማሲ | ዲኘሎማሲ |
| ዲሲፕሊን | ዲሲኘሊን |
| ጎንዳር | ጐንዳር |
| ጎጃም | ጐጃም |
| ጕበት | ጉበት |
| ጕዳይ | ጉዳይ |
| ፕሪሚየር | ኘሪሚየር |
| ፕራይቬታይዜሽን | ኘራይቬታይዜሽን |
| ፕሬስ | ኘሬስ |
| ፕሬዚዳንት | ኘሬዚዳንት |
| ፕሮጀክት | ኘሮጀክት |
| ፕሮግራም | ኘሮግራም |
| ፕሮፌሰር | ኘሮፌሰር |
| ፕሮፖጋንዳ | ኘሮፖጋንዳ, ፕሮፓጋንዳ, ኘሮፓጋንዳ |
| ፖሊቲካ | ፓሊቲካ |

### Type 3: False Ge’ezisms

| Canonical Amharic | Common Amharic |
|---|---|
| ሀገር | ኃገር |
| ሁለት | ኍለት, ኹለት |
| ሁሉ | ኍሉ, ኹሉ |
| ምልክት | ምልእክት, ምልዕክት |
| ትዕግሥት | ትዕግዕሥት |

## Type 5: Orthographic Abbreviations and Elisions

| Canonical Amharic | Common Amharic |
|---|---|
| መልአኩ | መላኩ |
| ሚያዝያ | ሚያዚያ |
| ማርያም | ማሪያም |
| ምክንያት | ምክኒያት |
| በንደ | በእንደ |
| ሰማንያ | ሰማኒያ |
| ነጉሣውያን | ነጉሣዊያን |
| አንባብያን | አንባቢያን |
| አብዮታውያን | አብዮታዊያን |
| ኢትዮጵያውያን | ኢትዮጵያዊያን |
| አንባብያን | አንባቢያን |
| ክርስቲያን | ክርስትያን |
| ወንጌላውያን | ወንጌላዊያን |
| የአማርኛ | ያማርኛ |

## Type 6: Disjoint Labiovelars

| Canonical Amharic | Common Amharic |
|---|---|
| ሆኗል | ሆኖዋል, ሆኖአል |
| ቍንጫ | ቁንጫ |
| ተቃውሞዋቸ | ተቃውሞአቸ |
| በእርስዋም | በእርሷም |
| ዓድዋ | ዓዷ, አድዋ |
| ይዟል | ይዞአል |
| ጆሮአቸውን | ጆሮዋቸውን |
| ገልጿል | ገልፀዋል |
| ጐረመሰ | ጎረመሳ |
| ጡዋት | ጥዋት, ጠዋት, ጧት |

## Type 7: Dialect Variations

| Canonical Amharic | Common Amharic |
|---|---|
| ሂጅ | ሂጂ |
| አይዶለም | አይደለም |
| ዐመፀ | ዐመጠ |
| ዓፄ | ዓጤ, አፄ, ሐፄ |

## Type 8: Foreign Language Transcription

| Canonical Amharic | Common Amharic |
|---|---|
| ቴክኖሎጂ | ቴክኒዎሎጂ |
| አቪኖር | አቢኖር |
| ኢሜይል | ኢሜል, ኤሜል, ኤሜይል |
| ኢንተርኔት | ኢንተርነት, ኢንቴርኔት, ኢንቴርኔት |
| ኮምፒዩተር | ኮምፒውተር |
| ፕሬዚዳንት | ፕረዚደንት |

## Type 9: Mistrikes

| Canonical Amharic | Common Amharic |
|---|---|
| ሥርዓት | ሥርአት, ሥራት |
| አርአያ | አራያ |
| ኢትዮጵያ | ኢትዮፕያ |
| ጤና | ጠና, ቴና, ጤኛ |
| ወጤት | ወጤጥ |